\documentclass[10pt]{article}

\usepackage{booktabs} 

\usepackage{geometry}
\usepackage{multicol}
\usepackage{float}
\usepackage[ruled,norelsize]{algorithm2e}
\usepackage[noend]{algorithmic}
\usepackage{calc}
\newcommand{\STATEWITHCOMMENT}[2]{\STATE{\makebox[\widthof{bla bla bla bla bla bla bla}][l]{#1}$\triangleright$ #2}}
\usepackage{subfig}
\usepackage{tikz}
\usepackage{bbm}
\usepackage{paralist}
\usepackage{enumitem}
\usepackage{color}
\usepackage{graphicx}
\usepackage{psfrag}
\usepackage{abstract}

\usepackage{amsthm}

\usepackage[normalem]{ulem}

\usepackage{amssymb}
\usepackage{amsmath}
\usepackage{hyperref}
\usepackage[utf8]{inputenc}

\usepackage{scalerel}

\allowdisplaybreaks

\newcommand{\erratum}[1]{{\color{blue}#1}}

\newcommand{\eone}{e_1}

\newcommand{\R}{\mathbb{R}}
\newcommand{\Normal}{\mathcal{N}}
\newcommand{\Expectation}{\mathbb{E}}
\newcommand{\E}{\mathbb{E}}

\newcommand{\id}{\mathrm{I}}

\newcommand{\ind}[1]{\mathbf{1}_{\{#1\}}}
\newcommand{\indlr}[1]{\mathbf{1}_{\left\{#1\right\}}}

\definecolor{dkgreen}{rgb}{0,0.6,0}
\definecolor{dkyellow}{rgb}{0.6,0.6,0}
\definecolor{dkred}{rgb}{0.6,0,0}
\definecolor{dkcyan}{rgb}{0,0.4,0.4}
\newcommand{\todo}[1]{{\color{red}~\\TODO:~#1}}
\newcommand{\yohe}[1]{{\color{blue}[Youhei]~#1}}
\newcommand{\anne}[1]{{\color{magenta}[Anne]~#1}}
\newcommand{\tobi}[1]{{\color{dkgreen}[Tobias]~#1}}

\newcommand{\bmat}{\begin{pmatrix}}
\newcommand{\emat}{\end{pmatrix}}
\newcommand{\F}{\mathcal{F}}
\newcommand{\barsigma}{\bar{\sigma}}

\newcommand{\note}[1]{{\color{magenta}{\bf Note} #1}}

\newcommand{\del}[1]{{\color{dkyellow}Del: {#1}}}

\newcommand{\uu}{u}
\newcommand{\elll}{\ell}

\newcommand{\target}{\beta}

\newcommand{\Y}{Y^{\scaleobj{0.7}{A}}}

\newtheorem{theorem}{Theorem}
\newtheorem{lemma}{Lemma}
\newtheorem{proposition}{Proposition}

\newtheorem{remark}{Remark}

\renewcommand{\del}[1]{}
\renewcommand{\todo}[1]{}
\renewcommand{\yohe}[1]{}
\renewcommand{\anne}[1]{}
\renewcommand{\tobi}[1]{}
\renewcommand{\note}[1]{}

\title{\textbf{
	Drift Theory in Continuous Search Spaces:\\
	Expected Hitting Time of the (1+1)-ES\\
	with 1/5 Success Rule
}}

\author{
	Youhei Akimoto\\
	Faculty of Engineering, Information and Systems, University of Tsukuba\\
	1-1-1 Tennodai, Tsukuba, Japan\\
	\texttt{akimoto@cs.tsukuba.ac.jp}
	\and
	Anne Auger\\
	RandOpt Team, INRIA Saclay; CMAP, Ecole Polytechnique\\
	Route de Saclay, Ile-de-France, France\\
	\texttt{anne.auger@inria.fr}
	\and
	Tobias Glasmachers\\
	Institute for Neural Computation, Ruhr-University Bochum\\
	Universitätsstr.~150, Bochum, Germany\\
	\texttt{tobias.glasmachers@ini.rub.de}
}

\date{}

\begin{document}

\maketitle

\setlength{\absleftindent}{4cm}
\setlength{\absrightindent}{4cm}

\begin{abstract}
\normalsize
This paper explores the use of the standard approach for proving runtime
bounds in discrete domains---often referred to as drift analysis---in
the context of optimization on a continuous domain. Using this framework
we analyze the (1+1) Evolution Strategy with one-fifth success rule on
the sphere function. To deal with potential functions that are not
lower-bounded, we formulate novel drift theorems. We then use the
theorems to prove bounds on the expected hitting time to reach a certain
target fitness in finite dimension $d$. The bounds are akin to linear
convergence. We then study the dependency of the different terms on $d$
proving a convergence rate dependency of $\Theta(1/d)$. Our results
constitute the first non-asymptotic analysis for the algorithm
considered as well as the first explicit application of drift analysis
to a randomized search heuristic with continuous domain.
\end{abstract}

\section*{Erratum}
The version of this paper published at ACM-GECCO 2018 contains some
technical errors. The errors do not affect the correctness of the
theorems. They are corrected in this version. For clarity, the changes
compared to the GECCO version are marked in \erratum{blue}.

\pagebreak

\begin{multicols}{2}

\section{Introduction}

The standard methodology for proving runtime bounds of evolutionary
algorithms defined on a discrete search space is often referred to as
drift analysis. It consists in proving a drift condition, e.g., expected
change strictly smaller than $c < 0$ (additive drift) w.r.t.\ a
potential, that directly translates into a bound on the hitting time to
reach the optimum. It allows to decouple generic mathematical arguments,
summarized in drift theorems, from arguments specific to the algorithm.
With drift analysis, proofs that could take several pages before have
been simplified considerably
\cite{Lehre12,Droste2002,Doerr:2012,lehre2013general,lengler2016drift}.

In this work we explore the utility of such an approach for the analysis
of algorithms operating on a continuous domain. For this purpose we
focus on the analysis of the $(1+1)$-ES with one-fifth success rule on
the sphere function $f : \R^d \to \R$, $x \mapsto \|x\|^2$. We are
particularly interested in benefits of drift analysis over current tools
for analyzing continuous randomized search heuristics, like
investigating stability of Markov chains.

\paragraph{The (1+1)-ES}
We focus here on one of the simplest adaptive algorithms, namely the
$(1+1)$ evolution strategy (ES) with one-fifth success rule
\cite{rechenberg:1973}. It is defined in
algorithm~\ref{algo}, where we assume \emph{minimization} of a function
$f:\R^d \to \R$. The state of the algorithm at iteration $t$ is
$(m_t, \sigma_t) \in \R^d \times \R_{>0}$, where $m_t$ is the mean of
the Gaussian sampling distribution and also the best solution found so
far, and $\sigma_t$ is the standard deviation of the distribution or
``step-size'' that controls the distance at which novel solutions are
sampled. This variant of the algorithm, which was first proposed in
\cite{kern:2004}, implements Rechenberg's idea of maintaining a
probability of success of roughly $1/5$.
This algorithm is not a ``toy'' algorithm as it features the important
flavor of the widely used state-of-the-art CMA-ES \cite{hansen:2001},
namely adaptation of the sampling distribution.

\begin{algorithm}[H]
\caption{(1+1)-ES with $1/5$-success rule}
\label{algo}
\begin{algorithmic}[1]
\STATE{\textbf{input} $m_0 \in \mathbb{R}^d$, $\sigma_0 > 0$, $f: \R^d \to \R$}, \textbf{parameter} $\alpha > {\color{blue}{1}}$
\FOR {$t = 1,2,\dots$, \textit{until stopping criterion is met}}
	\STATE {sample $x_t \sim m_t + \sigma_t \Normal(0,  I)$}
	\IF {$f\big(x_t\big) \leq f\big(m_t\big)$}  
		\STATEWITHCOMMENT{$m_{t+1} \leftarrow x_t$}{move to the better solution}
		\STATEWITHCOMMENT{$\sigma_{t+1} \leftarrow \sigma_t \cdot {\color{blue} \alpha}$}{increase the step size}
	\ELSE 
		\STATEWITHCOMMENT{$m_{t+1} \leftarrow m_t$}{stay where we are}
		\STATEWITHCOMMENT{$\sigma_{t+1} \leftarrow \sigma_t \cdot {\color{blue} \alpha^{-1/4}}$}{decrease the step size}
	\ENDIF
\ENDFOR
\end{algorithmic}
\end{algorithm}

\paragraph{Drift Analysis in $\R^d$}
Interestingly, although drift theorems are often formulated for finite
domains, they naturally generalize to continuous domains \cite{lehre2013general,lengler2016drift}.
To date however, drift analysis in the style of discrete domains
has not been \emph{explicitly} applied to analyze continuous algorithms.
Note that drift conditions are also central in other approaches
addressing convergence in continuous domains, while they are typically
not used for obtaining bounds on the hitting time (see below).
At the same time, we will see that some difficulties can arise when
dealing with continuous search spaces as it seems natural to use a
potential function that converges to minus infinity when approaching the
optimum. To overcome those problems we formulate novel drift theorems.

Analyzing state-of-the art continuous evolutionary algorithms means
analyzing adaptive algorithms. While this adaptation is the key for the
practical success of ES (ensuring linear convergence on wide classes of
problems, similar to gradient-based methods on strongly convex
functions), in turn it makes the analysis difficult. Indeed, when
$\sigma_t$ is too small compared to $\|m_t\|$, the progress towards the
optimum is very small. This complicates the task of finding a suitable
potential function and proving a drift condition.

When analyzing algorithms in continuous domains, our goals are (i) to
establish how fast the algorithm converges for a fixed dimension~$d$
(usually linear convergence), and (ii) to investigate the dependency of
the convergence rate on the search space dimension (usually
$\Theta(1/d)$)---this is different from discrete domains where the
optimum can be located in finite time. In terms of hitting time to reach
a certain precision $\epsilon$, property (i) means that for all
$\epsilon > 0$ the expected hitting time is finite and proportional to
$\log(\|m_0\|) - \log(\epsilon)$, while property (ii) means that it is
also proportional to~$d$.

\paragraph{Related work}
Some of the drift methodology is underlying many results of J.\ Jägersküpper
\cite{jaegerskuepper2003analysis,jagerskupper2006quadratic,jagerskupper2007algorithmic}.
Drift is not uncovered explicitly in these works, which makes it
arguably difficult to follow the analysis carried out. That might be the
reason why nobody built so far on Jägersküpper's impressive work.
We also have to point out that Algorithm~\ref{algo} differs from the
variant analyzed by Jägersküpper, where the step-size is kept fixed for
several iterations.

For a fixed dimension, the linear convergence of the algorithm on
scaling-invariant functions---including in particular the sphere
function---has been shown using Markov chain analysis \cite{AugerH13a}.
This analysis is asymptotic in nature and does not provide a dependency
of the convergence rate on the dimension. The difficult part in the
approach also relies on proving a drift condition, that should however
hold only outside a compact set, not on the whole domain.

Drift of a step size adaptive algorithm is also analyzed in
\cite{correa2016lyapunov}, the only prior work that uses a potential
function in a continuous domain. That approach
remains very restricted, applying only to symmetric functions of
a single variable.

In this work, we go beyond the state-of-the-art as follows. Other than
Jägersküpper's results, our bounds provide (non-asymptotic) constants,
and they hold with full probability. In contrast to Markov chain
analysis, we obtain a dependency in the dimension and non-asymptotic
results. Compared to \cite{correa2016lyapunov}, we go beyond a proof of
concept by analyzing a simple yet realistic algorithm.

\paragraph{Outline}
The rest of the paper is organized as follows. In the next section
we introduce novel drift theorems for lower and upper bounds to deal with
unbounded potentials, since this is a natural design in our context. In
Section~\ref{sec:technical}, we prove technical results needed to derive
the drift condition for the upper bound. In Section~\ref{sec:potential-drift}
we define our potential function and show two drift conditions for the
lower and the upper bound. By applying the drift theorems to the drift
conditions we derive lower and upper bounds on the first hitting time,
corresponding to linear convergence with $\Theta(1/d)$ scaling of the
convergence rate.
For the sake of readability, all proofs are in the appendix.

\paragraph{Notation}
A multivariate normal distribution is denoted $\Normal(0, I)$.
With $\Phi_1$ we denote the cumulative density function of the standard
normal distribution $\Normal(0, 1)$ on $\R$, and $\varphi_d$ is the pdf
of the standard normal distribution on $\R^d$.
\del{$\ind{C}$ denotes the indicator function of the set or condition~$C$.}
The indicator function of a set or condition $C$ is denoted by~$\ind{C}$.

\section{Additive Drift on an Unbounded Domain}
\label{section:drift}

In the continuous setting considered in this paper, we aim at proving
a runtime bound that translates into linear convergence. Linear
convergence is typically pictured as the log of the distance to the
optimum converging to minus infinity like $-{\rm CR} \times t$ with
${\rm CR} > 0$. It is thus natural to construct a potential function
that involves the log of the distance to the optimum. Yet, this means
that the potential function can take values that are arbitrarily
negative, while in drift theorems it is typical to assume that the
potential function is lower bounded (by zero or one). For this reason
we need to adapt existing drift theorems.

We adopt the following formalism. Let $\{ X_t : t \geq 0 \}$ be a
sequence of real-valued random variables adapted to a filtration
$\{\F_t : t \geq 0\}$. In our typical setting $X_t$ can be homogeneous
to the logarithm of the distance to the optimum and thus go to minus
infinity when linear convergence occurs. Additionally, from one
iteration to the next, $X_{t+1}$ can be arbitrarily much smaller than
$X_t$. This happens if by chance we have made an atypically good step
that improves the current solution a lot.

While arbitrarily good steps should be \emph{helpful} in the sense of
making the hitting time only smaller, we face the technical difficulty
to distinguish this situation from the following scenario: assume a
process $X_t$ with an average decrease of $-1$, i.e., fulfilling
$\E[ X_{t+1} | \F_t ] - X_t \leq -1$, but where $X_{t+1}$ equals $X_t$
with probability $1-p$, and with probability $p \ll 1$ we jump to
$X_{t-1} = X_t - 1/p$, possibly overjumping the target in a single but
very improbable step. The time needed to sample this jump is
geometrically distributed with expectation $1/p$, resulting in an
arbitrarily large hitting time. This small example illustrates that
controlling only the drift is not enough for bounding the expected
hitting time. If the domain is bounded from below then the size of a
possible jump is also bounded, avoiding this difficulty. Therefore we
have to find a way of controlling extreme events.

To circumvent this problem, instead of controlling directly the drift on
$X_t$ we will control the drift of a process with truncated and hence
bounded single-step progress. More precisely, for given $A>0$ we
consider the truncated process defined iteratively as $\Y_0 = X_0$ and
\begin{equation}
  \Y_{t+1} = \Y_t + \max \Big\{ X_{t+1} - X_t , -A \Big\}
  \enspace,
  \label{eq:truncatedprocess}
\end{equation}
where progress (towards minus infinity) larger than $-A$
is cut. By construction (almost surely%
  \footnote{We use \emph{almost surely} although the property is
  deterministic, simply to disambiguate from \emph{in distribution} and
  \emph{in expectation}.})
\begin{align}\label{eq:AS-bound}
& \Y_{t+1} - \Y_t \geq -A \enspace, \\\label{eq:XsmallerY}
& X_t \leq \Y_t \enspace,
\end{align}
where the latter equation holds since $X_t = X_0 + \sum_{k=0}^{t-1} (X_{k+1} - X_{k}) \leq \Y_0 + \sum_{k=0}^{t-1} \max \{ (X_{k+1} - X_{k} ), -A \} = \Y_0 + \sum_{k=0}^{t-1} (\Y_{k+1} - \Y_k) = \Y_t$.

As a direct consequence of inequality~\eqref{eq:XsmallerY},
for $\target \in \R$, the hitting time
$T_\target^X = \min \{ t : X_t \leq \target \} \in \mathbb{N} \cup \{\infty\}$
of $X_t$ to reach $(-\infty, \target]$ is upper bounded by the hitting
time $T_\target^{\Y} = \min \{ t : \Y_t \leq \target \}$ of $\Y_t$ to
reach $(-\infty, \target]$, i.e., $T_\target^X \leq T_\target^{\Y}$.
Hence an upper bound on the hitting time of $\Y_t$ results in an upper
bound on the hitting time of $X_t$. Exploiting this idea, we derive an
upper bound on the hitting time $T^{X}_\target$ in the following theorem
based on bounding the drift of the truncated process
$\{\Y_t: t \in \mathbb{N} \}$.

\begin{theorem}[Upper bound via drift on truncated process]
\label{theo:drift-UB-trunc}
Let $\{ X_t : t \geq 0 \}$ be a sequence of real-valued random variables
adapted to a filtration $\{ \F_t : t \geq 0\}$ with $X_0 = x_0 \in \R$.
For $\target < x_0$ let $T^X_\target = \min \left\{ t : X_t \leq \target \right\}$
be the first hitting time of the set $(-\infty, \target]$.
If there exist $A, B > 0$ such that $\Y_t$ is
integrable, i.e.\ $\E\left[\big|\Y_t\big|\right] < \infty$, and
\begin{equation}\label{eq:drift-truncated}
	\E \Big[ \Y_{t+1} \,\Big|\, \F_t \Big] - \Y_t = \E \Big[ \max \big\{X_{t+1} - X_t, -A \big\} \,\Big|\, \F_t \Big] \leq - B
	\enspace,
\end{equation}
then the expectation of $T^X_\target$ satisfies
\begin{equation}\label{Bound-HittingTime}
	\E \left[ T^X_\target \right] \leq \E \left[ T^{\Y}_\target \right] \leq \frac{x_0 - \target + A}{B}
	\enspace.
\end{equation}
\end{theorem}
Under slight misuse of terminology we define the \emph{truncated drift}
as the expected truncated one-step change as in equation~\eqref{eq:drift-truncated}.
\begin{remark}
A drift on the truncated process $\Y_t$ also gives a drift on $X_t$.
Indeed, assume $\E[| X_t| ]< + \infty$. Since
$$
	X_{t+1} - X_t \leq \max \left\{ X_{t+1} - X_t, -A  \right\} = \Y_{t+1} - \Y_t \enspace,
$$
if inequality~\eqref{eq:drift-truncated} is satisfied then it holds
\begin{equation}\label{eq:drift-unbounded}
\E[  X_{t+1} | \F_t ] - X_t \leq - B \enspace.
\end{equation}
\end{remark}

The next proposition ensures that the integrability of the truncated
process is implied by the integrability of $\{X_t : t \geq 0\}$.
\begin{proposition}[Integrability of the truncated process]\label{prop:integrability}
If a process $\{X_t : t \geq 0\}$ is integrable, i.e., $\E[|X_t|] < \infty$,
then its truncated process $\{\Y_t : t \geq 0\}$ defined in
equation~\eqref{eq:truncatedprocess} is integrable as well.
\end{proposition}

Our lower bound also relies on an unbounded potential function.
Typical drift theorems for establishing lower bounds assume that the
potential is bounded and hence cannot be applied directly
\cite{jagerskupper2007algorithmic}. Instead we use the following
theorem, the proof of which can be seen as a reformulation of the
arguments used in \cite[Theorem~2]{jagerskupper2006quadratic}
as a drift theorem. It generalizes
\cite[Lemma~12]{jagerskupper2007algorithmic}. Note that due to the more
general setting we lose a (bearable) factor of four in the bound.
\begin{theorem}
\label{theorem:drift-anne-jens}
Let $X_t$ be integrable and adapted to $\F_t$ such that
$$
	X_0 = x_0, \quad X_{t+1} \leq X_t, \quad\text{and}\quad \E[ X_{t+1} \,|\, \F_t ] - X_t \geq -C
$$
for $C > 0$. For $\target < x_0$ we define
$T^X_\target = \min \left\{ t : X_t \leq \target \right\}$.
Then the expected hitting time is lower bounded by
$$
	\E \left[ T_\target^X \right] \geq \frac{x_0 - \target}{4 C} - \frac12
	\enspace.
$$
\end{theorem}

\section{Probability of Successes with Positive Progress Rate}
\label{sec:technical}

\begin{figure}[H]
\begin{center}
\includegraphics[width=\columnwidth]{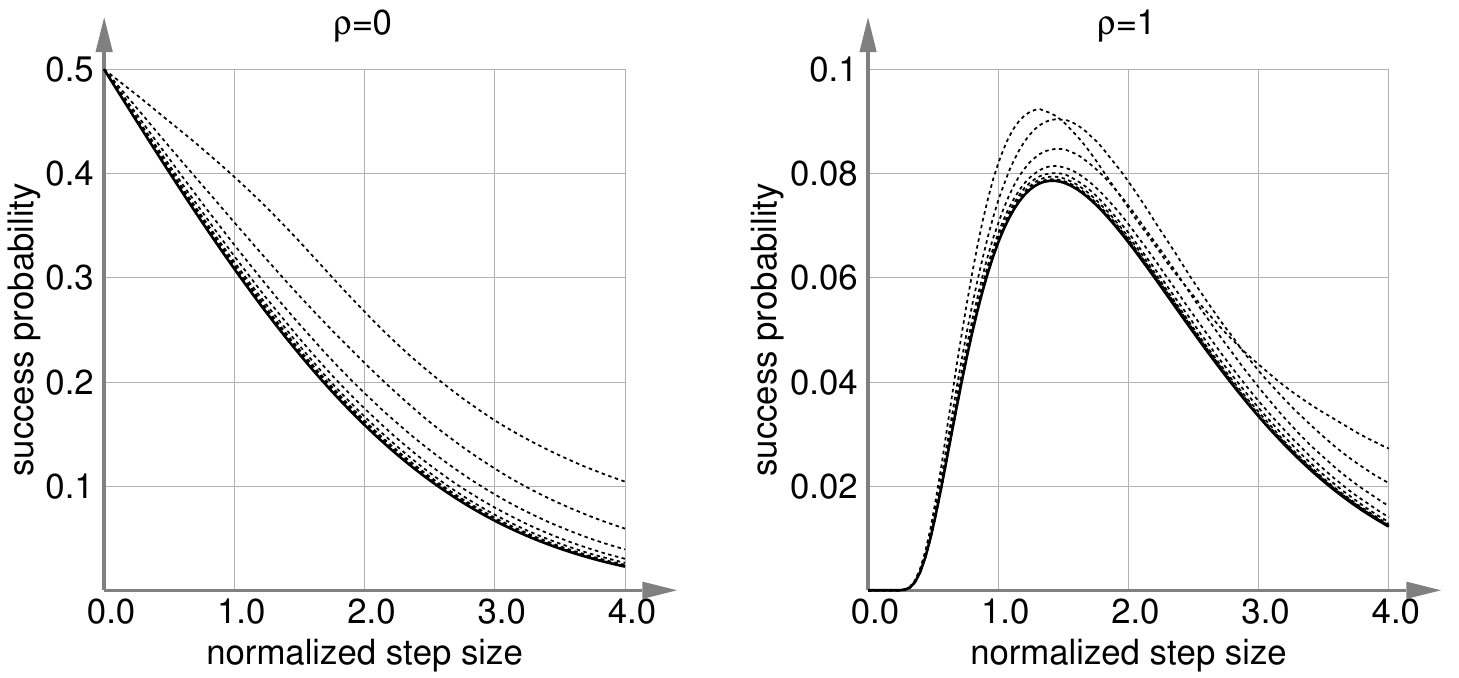}
\end{center}
\caption{
	The success probability function $p^\text{succ}_{d,r}(\barsigma)$
	for $r \cdot d = \rho = 0$ (left) and $r \cdot d = \rho = 1$ (right).
	The solid curves depict $p^\text{succ}_{\rho}(\barsigma)$, while the
	dotted curves are $p^\text{succ}_{r, d}(\barsigma)$ with
	$r \cdot d = \rho$ for $d \in \{2, 4, 8, 16, 32, 64, 128, 256\}$.
	The curves for high dimensions are indistinguishable from the limit
	curve.
	\label{figure:success2D}
}
\end{figure}

In this section we derive properties of the success probability that
will be central for establishing the drift condition for the upper
bound.
For an improvement rate $r \in [0, 1)$ and
$x \sim \Normal(m, \sigma^2 \id)$ in $\R^d$ we define the success
probability with rate $r$ given $(m,\sigma)$ as
\begin{align*}
p^\text{succ}_{r,d}(m, \sigma) = \Pr_{x \sim \Normal(m, \sigma^2 \id)}\Big(\|x\| < (1-r) \cdot \|m\|\Big)
\end{align*}
i.e.\ as the probability that the norm of the offspring is smaller than
$(1-r) \| m \|$. As a consequence of the isotropy of the multivariate
normal distribution, this success probability equals
$$
p^\text{succ}_{r,d}(m, \sigma) = \Pr \left( \left\| \eone + \frac{\sigma}{\|m\|} \Normal \right\| < (1-r) \right)
$$
where $\eone=(1,0,\ldots,0)$ and $\Normal$ is a standard normally
distributed vector. This latter equation reveals that the probability of
success with improvement rate $r$ is a function of $\sigma/\|m\|$. Let
us introduce the normalized step size
$\barsigma = d \cdot \sigma / \| m \|$ and define
\begin{equation}\label{eq:psuccr}
	p^\text{succ}_{r,d}(\barsigma) : =  \Pr  \left( \left\| \eone + \frac{\barsigma}{d} \Normal \right\| < (1-r) \right) \enspace,
\end{equation}
then $p^\text{succ}_{r,d}(\barsigma) = p^\text{succ}_{r,d}(m, \sigma)$.
For $r=0$ we recover the ``classic'' probability of success
$$
p^\text{succ}_{0,d}(\barsigma) := \Pr \left( \left\| \eone + \frac{\barsigma}{d} \Normal \right\| < 1 \right) \enspace.
$$
The success probability function is illustrated in Figure~\ref{figure:success2D}.
We start by proving that the function
$\barsigma \to  p^\text{succ}_{r,d}(\barsigma)$ is continuous, and for
$r=0$ it is monotonically decreasing and hence bijective.
This is formalized in the following lemma:
\begin{lemma}\label{lemma:bijection}
\begin{enumerate}
\item
	For all $d \in \mathbb{N}$ and $r \in [0, 1)$,
	$\barsigma \to p^\text{succ}_{r,d}(\barsigma)$ is positive and continuous.
\item
	For $r=0$ it is strictly monotonically decreasing and thus bijective.
\item
	For all $d \in \mathbb{N}$, the image of $\barsigma \to p^\text{succ}_{0,d}(\barsigma)$ is
	$(0, 1/2)$.
	\end{enumerate}
\end{lemma}
We now investigate the asymptotic limit of the function
$\barsigma \to p^\text{succ}_{r,d}(\barsigma)$ for $d$ to infinity.
\begin{lemma}
\label{lemma:rho}
	For $r = r(d)$ fulfilling $\lim_{d \to \infty} d \cdot r(d) = \rho$
	the limit
	$p^\text{succ}_\rho(\barsigma) := \lim_{d \to \infty} p^\text{succ}_{r,d}(\barsigma)$
	exists, and it equals
	$p^\text{succ}_\rho(\barsigma) = \Phi_1\left(-\frac{\rho}{\barsigma} - \frac{\barsigma}{2}\right)$.
	For $\rho = 0$, the function $p^\text{succ}_0$ is
	continuous and strictly monotonically decreasing and the image of
	$p^\text{succ}_0$ is $(0, 1/2)$.
\end{lemma}

For $\rho=0$ we recover the known result that the asymptotic limit of
the probability of success (for $r=0$) equals $\Phi_1(-\sigma/2)$
\cite{Auger-FOGA-2011}.
The above lemma captures the intuition that success is maximized with a
small step size (for $\rho = 0$, $p^\text{succ}_\rho$ is maximal for
${\barsigma} \to 0$), while a non-trivial step size (${\barsigma} > 0$)
is needed for making significant progress ($\rho > 0$).

\section{Potential and Drift}\label{sec:potential-drift}

In this section we define a potential function $V(\theta_t)$ that
gives rise to the unbounded and untruncated process from
Section~\ref{section:drift}. First we establish that it satisfies the
conditions of Theorem~\ref{theo:drift-UB-trunc}. Then we prove a drift
condition for the lower bound. Finally we apply the drift theorems to
obtain lower and upper bounds for the first hitting time of the (1+1)-ES.
Our goal is to establish lower and upper bounds on the expected first
hitting time of $\log(\|m_t\|)$ to the set $(-\infty, \target]$, where
$\target = \log(\epsilon)$ is the logarithm of the target distance
$\epsilon$ to the optimum.
Linear or geometric convergence of (1+1)-ES---that is what we observe in
simulation and what Jägersküpper found in his analysis with
overwhelming probability---is implied if $\log(\|m_t\|)$ decreases at a
linear rate towards $-\infty$. The potential function $V(\theta_t)$ will
be chosen so that its first hitting time gives an upper bound on the
first hitting time of $\log(\|m_t\|)$.

\subsection{Potential Function}

We fix two probabilities $p_u$ and $p_{\ell}$ such that
$0 < p_u < 1/5 < p_{\ell} < 1/2$. Since the probability of success
function $\barsigma \mapsto p^\text{succ}_{0,d}(\barsigma)$ with rate
$r = 0$ is bijective (see Lemma~\ref{lemma:bijection}), we know that
there exist $\uu$ and $\elll$ such that $p^\text{succ}_{0,d}(\uu) = p_u$
and $p^\text{succ}_{0,d}(\elll) = p_{\ell}$.
We assume that $p_u$ and $p_\ell$ are chosen such that
$\uu / \elll \geq \alpha^{5/4}$. Given these parameters, we define the
potential function
\begin{align}
	&V(\theta) = V(m, \sigma) = \log \big( \|m\| \big) \label{eq:potential}
	\\
	&+ \max \Bigg\{
		0
		\, , \,\,
		v \cdot \log \left( \frac{\alpha \cdot \elll \cdot \|m\|}{d \cdot \sigma} \right)
		\, , \,\, 
		v \cdot \log \left( \frac{\alpha^{1/4} \cdot \sigma \cdot d}{\uu \cdot \|m\|} \right)
	\Bigg\}
	\notag
\end{align}
with coefficient $v > 0$ to be determined later.
The potential function consists of three parts. The term $\log(\|m\|)$
measures optimization progress: when approaching the optimum, it decays
to $-\infty$. The other terms become positive and hence active only if
the step size is not well adapted. The second term in the maximum kicks
in if $\sigma$ is ``too small'', and the third term turns positive if
$\sigma$ becomes ``too large''. Hence the potential combines two ways of
making progress, namely approaching the optimum and adapting the step
size towards a regime where the (1+1)-ES can make significant
optimization progress. The parameter $v$ relates these two types of
progress by putting them on the same scale.

\begin{lemma}
\label{lemma:integrable}
It holds $\Expectation\big[|V(\theta_t)|\big] < \infty$.
In other words, $V(\theta_t)$ is integrable for each $t \in \mathbb{N}$.
Moreover, for all $A > 0$ the truncated process $\Y_{t}$
defined in equation~\eqref{eq:truncatedprocess} with $X_t = V(\theta_t)$
is integrable for each $t \in \mathbb{N}$.
\end{lemma}

\subsection{Truncated Drift}

In the following, we prove that $V(\theta_t)$ satisfies the
prerequisites of Theorem~\ref{theo:drift-UB-trunc}. First we prove a
proposition with a range of possible choices for the constants
$A$ and $v$. We then show in Proposition~\ref{proposition:scaling}
how to set those constants to obtain the right scaling with respect to
$d$ for the hitting time.

\begin{proposition}
\label{proposition:drift}
Consider optimization of the sphere function $f : \R^d \to \R$,
$x \mapsto \|x\|^2$ with the (1+1)-ES.
If the parameters $v$ and $A$ fulfill
$0 < v < \min \left\{ 1, \erratum{A / \log(\alpha)} \right\}$ 
then the potential function $V$ defined in eq.~\eqref{eq:potential}
fulfills
\begin{equation}
	\Expectation\left[ \max\{ V(\theta_{t+1}) - V(\theta_{t}) \, ,\, -A \} \mid \theta_{t}\right]  \leq -B
	\enspace,
	\label{eq:boundopt}
\end{equation}
with
\begin{align}
B = & \min\Bigg\{ A \cdot p^* - \frac54 \cdot v \cdot \log (\alpha), \notag \\
	& v \cdot \log(\alpha) \cdot \left( \frac{5 p_{\ell} - 1}{4} \right),
	  v \cdot \log(\alpha) \cdot \left(\frac{1 - 5  p_{u}}{4} \right) \Bigg\}
	\label{eq:B}
\end{align}
and
$p^* = \min\limits_{\barsigma \in [\elll, \uu]} \left\{ p^\text{succ}_{1 - \exp\left( - \frac{A}{1 - v}\right), d} (\barsigma) \right\} $.
\end{proposition}

The previous proposition is the core component establishing the drift of
the truncated process. The next proposition shows how to arrange the
parameters so that the speed of the drift scales as desired in the limit
of large dimensions.

\begin{proposition}
\label{proposition:scaling}
Consider $d \geq 2$.
For $A = \frac{1}{d}$ and
$v = \frac{p'}{2 \cdot d \cdot \log(\alpha)}$
with
$p' = \min\limits_{\barsigma \in [\elll, \uu]} \left\{ p^\text{succ}_{r', d}(\barsigma) \right\}$
and
$r' = 1 - \exp\left(-\frac{\log(\alpha)}{d \cdot \log(\alpha) - 1}\right)$
it holds $B > 0$ and $B \in \Theta(1/d)$.
\end{proposition}

Proposition~\ref{proposition:scaling} implies that the truncated
drift~\eqref{eq:boundopt} is of order~$\Omega(1/d)$.

\subsection{Hit-and-Run}

A very general lower bound on the expected first hitting time was
established by Jägersküpper. His argumentation in
\cite[Theorem~2]{jagerskupper2006quadratic} is based on the
hit-and-run algorithm. Here we use a similar approach for proving the
lower bound.
In iteration $t$, given a mutation direction $\delta_t = x_t - m_t$
(with the notation of algorithm~\ref{algo}), the hit-and-run algorithm
selects the optimal length of $\delta_t$ maintaining its direction and
produces the offspring $x^*_t = m_t + \gamma^* \delta_t$ with
$\gamma^* = \arg\min_\gamma f(m_t + \gamma \cdot \delta_t)$.
By construction, the progress of the hit-and-run method upper bounds the
progress of the (1+1)-ES. Using the same realization for the Gaussian
vector creating the offspring $x_t$ (see Algorithm~\ref{algo}), we
indeed have:
\begin{equation}\label{hit-and-run-boundingopo}
	\log(\| m_t \|) - \log(\| m_{t+1} \| )  \leq \log(\| m_t \|) - \log(\| x^*_t \| )
	\enspace.
\end{equation}
The log-progress of the hit-and-run on the sphere is 
$$
	\log\big(\|m_t\|\big) - \log\big(\|x^*_t\|\big) = - \log \Big(\Big\| e_1 + (\gamma^* / \|m_t\|) \delta_t\Big\|\Big)
	\enspace.
$$
In the next lemma, we bound the expectation of its progress.
\begin{lemma}\label{lemma:hitandrun}
For $d \geq 2$, the expected log progress of the hit-and-run algorithm
is upper bounded by $1/d$.
\end{lemma}
Using inequality~\eqref{hit-and-run-boundingopo} we find that the
expected log progress of the (1+1)-ES is upper-bounded by $1/d$:
\begin{equation}\label{LB-drift}
\E \big[ \log(\|m_t\|) - \log(\|m_{t+1}\|) \,\big|\, \F_t \big] \leq \frac{1}{d}.
\end{equation}

\subsection{Bounds on the First Hitting Time}

Finally, all preparations are in place and we can reap the fruit of our
labor, which are formulated in the following theorem.
To this end, let $T_\epsilon = \min \{ t : \| m_t \| \leq \epsilon \}$
be the first hitting time of $(-\infty, \log(\epsilon)]$ by $\log(\|m_t\|)$,
where $m_t$ is defined in Algorithm~\ref{algo}.

\begin{theorem}
\label{theorem:main}
The expected first hitting time of the (1+1)-ES starting from
$\theta_0 = (m_0, \sigma_0)$ on the sphere function $f(x) = \|x\|^2$
fulfills
$$
	\frac{\big(\log(\|m_0\|) - \log(\epsilon)\big) \cdot d}{4} - \frac{1}{2}
	\leq \E[T_\epsilon]
	\leq \frac{V(\theta_0) - \log(\epsilon) + \frac1d}{B}
$$
with $V(\theta)$ defined in eq.~\eqref{eq:potential} and
$B$ given in eq.~\eqref{eq:B}. With the choice of constants $A$
and $v$ given in Proposition~\ref{proposition:scaling}, it is hence of the form
\begin{align}
	\E[T_\epsilon] \in \Theta\Big(\big(\log(\|m_0\|) + \log(1/\epsilon)\big) \cdot d\Big)
	\label{eq:asymptotics}
	\enspace.
\end{align}
\end{theorem}
The asymptotic form~\eqref{eq:asymptotics} of the expected first
hitting time implies (i) that the process is akin to linear convergence
due to the term $\log(1/\epsilon)$, and (ii) a convergence rate of the
form $\Theta(1/d)$ due to the factor $d$ in the expected hitting time.

\section{Discussion and Conclusion}
We have established the first non-asymptotic runtime bound for the first
hitting time of the (1+1)-ES with one-fifth success rule
(Algorithm~\ref{algo}) on the sphere function.
Our proof is based on a global drift condition,
a generic approach that has proven invaluable for the analysis of
discrete algorithms. Our work shows that
such approaches are a promising tool also for continuous domains.
As usual in drift analysis, constructing the potential function and
establishing drift conditions makes up the lion's share of the efforts.
In this sense, our drift theorems merely add convenience.

Establishing a drift condition is simplified in the stability analysis
of the underlying Markov chain, since drift is needed only outside a
compact set, i.e., for very small and very large normalized step size
$\barsigma$. On the other hand, the current analysis is non-asymptotic
and provides estimates of the convergence rate as a function of the
problem dimension.

Jägersküpper established similar results already more than a decade ago,
when drift analysis was only in its infancy.
His results are hard to follow from a modern perspective.
\del{Also, the results are asymptotic and don't provide bounds for finite dimension.}
We improve on his work by proving non-asymptotic bounds for finite dimensions.

\paragraph{\textbf{Acknowledgement}}
We gratefully acknowledge support by Dag\-stuhl seminar 17191
``Theory of Randomized Search Heuristics''. We would like to thank
Per Kristian Lehre, Carsten Witt, and Johannes Lengler for valuable
discussions and advice on drift theory.

{
\footnotesize
\bibliographystyle{plain}

}

\newpage
\section*{Appendix}

\begin{proof}[proof of theorem~\ref{theo:drift-UB-trunc}]
We consider the truncated process defined above 
and the stopped truncated process as $Z_0 = \Y_0$ and
$Z_{t} = \Y_{\min\{t, T^{\Y}_\target\}}$. By construction it holds
$X_t \leq \Y_t \leq Z_t$ and $T^{X}_\target \leq T^{\Y}_\target$.
We will prove that
\begin{equation}\label{eq:super-martingale}
	E[Z_{t+1} \,|\, \F_t ] \leq Z_t - B \cdot \ind{T^{\Y}_\target > t} \enspace.
\end{equation}
We start from
\begin{equation}\label{eq:add-two}
	E[Z_{t+1} \,|\, \F_t ] = E[ Z_{t+1} \ind{ T^{\Y}_\target \leq t } \,|\, \F_t ] + E[ Z_{t+1} \ind{ T^{\Y}_\target > t } \,|\, \F_t ]
\end{equation}
and estimate the different terms:
\begin{equation}\label{eq:top}
	E[ Z_{t+1} \ind{ T^{\Y}_\target \leq t } \,|\, \F_t ]  = E[ Z_{t} \ind{ T^{\Y}_\target \leq t } \,|\, \F_t ]  = Z_{t} \ind{ T^{\Y}_\target \leq t }
\end{equation}
where we have used that $\ind{ T^{\Y}_\target \leq t }$ is
$\F_t$-measurable, and this also implies that $\Y_t$ and $Z_t$, being
functions of $X_t$ and $T^{\Y}$, are $\F_t$-measurable.
\note{[To keep for ourselves: it is $\F_t$ measurable because we can say whether $T^{\Y}_\target \leq t$ by knowing all what happened till time $t$, i.e. $\ind{ T^{\Y}_\target \leq t }$ is a deterministic function of the random variables in the past till time $t$.]}
Also
\begin{multline}\label{eq:LEEXtt}
	E[ Z_{t+1} \ind{ T^{\Y}_\target > t } \,|\, \F_t ] = E[ Y_{t+1} \,|\, \F_t] \ind{ T^{\Y}_\target > t } \\ \leq (Y_t - B) \ind{ T^{\Y}_\target > t } = (Z_t - B) \ind{ T^{\Y}_\target > t }
\end{multline}
where we have also used that $\ind{ T^{\Y}_\target > t }$ is $\F_t$ measurable.
Hence injecting \eqref{eq:top} and \eqref{eq:LEEXtt} into \eqref{eq:add-two}, we end up with \eqref{eq:super-martingale}.
From \eqref{eq:super-martingale}, by taking the expectation we deduce
\begin{equation}
\E[ Z_{t+1}  ] \leq \E[ Z_{t}] - B \cdot \Pr[T_A > t] \enspace.
\end{equation}
Following the same approach as \cite[Theorem~1]{lengler2016drift}, since
$T^{\Y}_\target$ is a random variable taking values in $\mathbb{N}$, it can be
rewritten as $\E[T^{\Y}_\target] = \sum_{t=0}^{+\infty} \Pr[T^{\Y}_\target > t]$
and thus it holds
\begin{multline}
	B \cdot \E\left[T^{\Y}_\target\right]
	\stackrel{\tilde{t} \to \infty}{\longleftarrow} \sum_{t=0}^{\tilde{t}} B \cdot \Pr\left[T^{\Y}_\target > t\right]
	\leq \sum_{t=0}^{\tilde{t}} \Big( \E[Z_t] - \E[Z_{t+1}] \Big) \\
	\leq \E[Z_0] - \E[Z_{\tilde{t}}] = x_0 - \E[Z_{\tilde{t}}]
	\enspace.
\end{multline}
Since $Y_{t+1} \geq Y_t - A$, then $Y_{T^{\Y}_\target} \geq \target - A$ and
given that $Z_t \geq Y_{T^{\Y}_\target}$, we deduce that $E[ Z_{\tilde{t}} ] \geq \target - A$
for all $\tilde{t}$, which implies
$$
\E \left[ T^{\Y}_\target \right] \leq \frac{x_0 - \target + A }{B} \enspace.
$$
With $\E[T^{X}_\target] \leq \E[T^{\Y}_\target]$ this proves the upper bound.
\end{proof}
\vspace{-0.8em}\noindent\rule{\linewidth}{0.3pt}\vspace{-1em}
\begin{proof}[proof of proposition \ref{prop:integrability}]
From the definition of the truncated process \eqref{eq:truncatedprocess} we obtain $|\Y_{t+1}| \leq |\Y_t| + |X_{t+1} - X_t| + A$ which implies
\begin{align*}
\E[|\Y_{t+1}|] 
&\leq \E[|\Y_t|] + \E[|X_{t+1} - X_t|] + A\\
&\leq \E[|\Y_t|] + \E[|X_{t+1}|] + E[|X_t|] + A 
\enspace,
\end{align*}
where the second to fourth terms are finite. Since $\Y_0 = X_0$ is integrable, $\Y_t$ is integrable by induction.
\end{proof}
\hrule
\begin{proof}[proof of theorem \ref{theorem:drift-anne-jens}]
After $T= \lfloor  (x_0 - \target) /  (2 C) \rfloor $ iterations it holds
$\E[ x_0 - X_{T} ] \leq C \cdot T \leq (x_0 - \target)/2 $. From Markov's
inequality we conclude $\Pr(x_0 - X_T \geq x_0 - \target) \leq \frac{1}{2}$
and thus $\Pr(x_0 - X_T \leq x_0 - \target) \geq \frac{1}{2}$, which is
equivalent to $\Pr(T^X_\target \geq T) \geq \frac{1}{2}$. Applying the
Markov inequality once more we obtain
\begin{equation*}
  \E[T^X_\target] \geq \Pr\left(T^X_\target \geq T\right) \cdot T \geq T/2 \geq \frac{x_0 - \target}{4 C} - \frac12
  \enspace.
  \qedhere
\end{equation*}%
\end{proof}
\hrule
\begin{proof}[proof of lemma \ref{lemma:bijection}]
We introduce the sample $z \sim \Normal(0, \id)$ through
$z = (x - m) / \sigma$, or equivalently, $x = m + \sigma \cdot z$.
Defining $A_{r,d}(\barsigma) := B \left( \frac{d}{\barsigma} \frac{-m}{\|m\|}, \frac{d}{\barsigma} (1-r) \right)$,
we write the success rate in the form
\begin{align*}
	p^\text{succ}_{r,d}(\barsigma) &=
	\int_{{A_{r,d}(\barsigma)}}
	\varphi_d(z) \, dz.
\end{align*}
For increasing values of ${\barsigma}$ the ball-shaped integration area shrinks,
and in case of $r > 0$ it also moves away from the origin. Together with
the monotonicity of $\varphi_d$ w.r.t.\ $\|z\|$ this proves that
$p^\text{succ}_{r,d}$ is monotonically decreasing. Continuity of
$p^\text{succ}_{r,d}$ follows from the boundedness of $\varphi_d$, and
positivity from the fact that $A_{r,d}(\barsigma)$ is non-empty and
$\varphi_d$ is positive. This proves the first claim. For $r = 0$ the
balls are nested. This immediately proves the second claim. From
\begin{align*}
	\bigcap_{\barsigma > 0} A_{0,d}(\barsigma) = \emptyset
	\quad \text{and} \quad
	\bigcup_{\barsigma > 0} A_{0,d}(\barsigma) = \Big\{ z \in \R^d \,\Big|\, m^T z < 0 \Big\}
\end{align*}
we conclude $\lim_{\barsigma \to 0} p^\text{succ}_{0,d}(\barsigma) = 1/2$
and $\lim_{\barsigma \to \infty} p^\text{succ}_{0,d}(\barsigma) = 0$,
which proves the last claim.
\end{proof}
\hrule
\begin{proof}[proof of lemma \ref{lemma:rho}]
We consider the sequence of random variables
\begin{align*}
	J_d = \indlr{ \left\| \eone + \frac{\barsigma}{d} \Normal \right\|^2 < (1-r)^2 }
		&= \indlr{ 1 + 2 \frac{\barsigma}{d} \Normal_1 + \frac{\barsigma^2}{d^2} \| \Normal \|^2 < 1 - 2 r + r^2 } \\
		&= \indlr{ 2 {\barsigma} \Normal_1 + \frac{\barsigma^2}{d} \| \Normal \|^2 < - 2 r d + r^2 d }
\end{align*}
indexed by $d$. Here $\Normal$ denotes a standard normally distributed
vector in $\R^d$, and $\Normal_1$ is its first component.
Almost surely by the Law of Large Numbers, $\| \Normal \|^2 / d$
converges to $1$ such that when $d$ goes to infinity then it holds
\begin{align*}
	\lim_{d \to \infty} \indlr{ 2 {\barsigma} \Normal_1 + \frac{\barsigma^2}{d} \| \Normal \|^2 <  - 2 r d + r^2 d } 
		= \indlr{ 2 {\barsigma} \Normal_1 + \barsigma^2 < - 2 \rho }
		= \indlr{ \Normal_1 < -\frac{\rho}{\barsigma} - \frac{\barsigma}{2} }
\end{align*}
almost surely.
Since $p^\text{succ}_{d,r}(\barsigma) = \E \left[J_d \right]$ and
$J_d$ converges almost surely to
$\indlr{ \Normal_1 < -\frac{\rho}{\barsigma} - \frac{\barsigma}{2} }$
we need to prove the uniform integrability to ensure that the limit also
holds in expectation. However the uniform integrability is here obvious
since $\E \left[|J_d| \right] \leq 1$ for all $d$. Hence we have proven
\begin{multline*}
	\lim_{d \to \infty} \E \left[J_d \right]
	= \E \left[ \indlr{ \Normal_1 < -\frac{\rho}{\barsigma} - \frac{\barsigma}{2} } \right]
	= \\\Pr\left( \Normal_1  < -  \frac{\rho}{\barsigma} - \frac12 \barsigma \right)
	= \Phi_1\left[-  \frac{\rho}{\barsigma} - \frac12 \barsigma\right]
	\enspace.
	\qedhere
\end{multline*}
\end{proof}
\hrule
\begin{proof}[proof of lemma \ref{lemma:integrable}]
The statement holds trivially for $t = 0$, since the initial
condition is a constant.
The following elementary calculation shows that the pole of the
logarithm in the definition of~$V$ is not problematic. Let $B(0, 1)$
denote the open ball of radius one around the origin, then we have:
\begin{alignat*}{2}
	 &\int_{B(0, 1)} \log(\|z\|) \, dz
	&&= \int_0^1 \int_{S(0, r)} \log(\|z\|) \, dz \, dr \\
	=& \int_0^1 \left( \int_{S(0, r)} \, dz \right) \log(r) \, dr
	&&= \frac{2 \cdot \pi^{d/2}}{\Gamma(d/2)} \int_0^1 r^{d-1} \log(r) \, dr \\
	=& \frac{2 \cdot \pi^{d/2}}{\Gamma(d/2)} \cdot \left[ \frac{r^d (d \log(r) - 1)}{d^2} \right]_0^1
	&&= -\frac{2 \cdot \pi^{d/2}}{\Gamma(d/2) \cdot d^2},
\end{alignat*}
where $\Gamma$ denotes the Gamma function. Therefore
$\Expectation\big[|V(\theta_t)| \,\big|\, \theta_{t-1}\big] < \infty$
for all $t$, and the statement follows by induction. 
The integrability of the truncated process is straight-forward from the above statement and Proposition~\ref{prop:integrability}.
\end{proof}
\hrule
\begin{proof}[proof of proposition~\ref{proposition:drift}]
For the sake of simplicity we introduce 
$\log^+(x) = \log(x) \cdot \indlr{ x \geq 1 }$.
We rewrite the potential function as
\begin{align}
	V(m_t,\ \sigma_t) = & \log \left( \|m_t\| \right) \notag \\
	&+ v \cdot \log^+ \left( \frac{\alpha \cdot \elll \cdot \|m_t\| }{\sigma_t \cdot d}  \right) \label{eq:a} \\
	&+ v \cdot \log^+ \left( \frac{\sigma_t \cdot d}{\alpha^{-1/4} \cdot \uu \cdot \|m_t\|} \right) \label{eq:b} \enspace.
\end{align}
We want to estimate the conditional expectation 
\begin{equation}\label{eq:cond-expe}
\Expectation\left[ \max\{ V(\theta_{t+1}) - V(\theta_{t}) \, ,\, -A \} \mid \theta_{t}\right].
\end{equation}
We partition the possible values of $\theta_t$ into three sets:
first the set of $\theta_t$ such that $\sigma_t < \elll \cdot \|m_t\| / d$ ($\sigma_t$ is small),
second the set of $\theta_t$ such that $\sigma_t > \uu \cdot \|m_t\| / d$ ($\sigma_t$ is large),
and last the set of $\theta_t$ such that $\elll \cdot \|m_t\| /d \leq \sigma_t \leq \uu \cdot \|m_t\|/d$ (reasonable $\sigma_t$).
In the following, we bound eq.~\eqref{eq:cond-expe} for each of the
three cases and in the end our bound $B$ will equal the minimum
of the three bounds obtained for each case.

\textit{Reasonable $\sigma_t$ case:
$\frac{\|m_t\|}{d \sigma_t} \in \left[ \frac{1}{\uu}, \frac{1}{\elll} \right]$}.
The potential function at time $t+1$ can be written as
\begin{align*}
	& V(\theta_{t+1}) = \log \left( \|m_{t+1}\| \right) \\
		&+ v \cdot \log \left( \frac{\alpha \cdot \elll \cdot \|m_{t+1}\|}{d \cdot \sigma_{t+1}} \right) 1_{\left\{ \alpha \elll \|m_{t+1}\| > d \cdot \sigma_{t+1} \right\}}1_{\left\{ \sigma_{t+1} > \sigma_{t} \right\}} \\
		&+ v \cdot \log \left( \frac{\alpha \cdot \elll \cdot \|m_{t+1}\|}{d \cdot \sigma_{t+1}} \right) 1_{\left\{ \alpha \elll \|m_{t+1}\| > d \cdot \sigma_{t+1} \right\}}1_{\left\{ \sigma_{t+1} < \sigma_{t} \right\}} \\
		&+ v \cdot \log \left( \frac{d \cdot \alpha^{1/4} \cdot \sigma_{t+1}}{\uu \cdot \|m_{t+1}\|} \right) 1_{\left\{\alpha^{-1/4} \uu \|m_{t+1}\| < d \cdot \sigma_{t+1} \right\}} 1_{\left\{ \sigma_{t+1} > \sigma_{t} \right\}} \\
		&+ v \cdot \log \left( \frac{d \cdot \alpha^{1/4} \cdot \sigma_{t+1}}{\uu \cdot \|m_{t+1}\|} \right) 1_{\left\{\alpha^{-1/4} \uu \|m_{t+1}\| < d \cdot \sigma_{t+1} \right\}} 1_{\left\{ \sigma_{t+1} < \sigma_{t} \right\}}.
\end{align*}
In case of success, where thus $\ind{\sigma_{t+1} > \sigma_t}=1$,
we have $\|m_{t+1}\| / \sigma_{t+1} \allowbreak < \|m_{t}\| / (\alpha \sigma_{t}) \leq d/ (\alpha \elll)$,
implying that the conditions in the second term never hold at the same
time and thus the second term is always $0$. Similarly, in case of failure,
$\|m_{t+1}\| / \sigma_{t} = \|m_{t}\| / (\alpha^{-1/4} \sigma) \leq d/
(\alpha^{-1/4} \uu)$ and we find that the fifth term is always zero.
We rearrange the third and fourth term into
\begin{align*}
	(3^\text{rd}) &= v \cdot \log^+ \left( \frac{\alpha^{5/4} \cdot \elll \cdot \|m_{t}\|}{d \cdot \sigma_{t}} \right) \cdot 1_{\left\{ \sigma_{t+1} < \sigma_t \right\}} \enspace, \\
	(4^\text{th}) &= - v \cdot \Bigg[ \log \left( \frac{ \|m_{t+1}\| }{ \|m_{t}\| } \right) - \log \left( \frac{d \cdot \sigma_{t}}{\alpha^{-5/4} \cdot \uu \cdot \|m_{t}\|} \right) \Bigg] \\
	&\qquad \times 1_{\left\{\alpha^{-5/4} \uu \|m_{t+1}\| < d \cdot \sigma_{t} \right\}} 1_{\left\{ \sigma_{t+1} > \sigma_t \right\}} \enspace.
\end{align*}
Then, the one-step change $\Delta_{t} = V(\theta_{t+1}) - V(\theta_{t})$ is upper bounded by
\begin{align*}
	\Delta_{t} \leq & \left(1 - v \cdot 1_{\left\{\alpha^{-5/4} \uu \|m_{t}\| < d \cdot \sigma_{t} \right\}} \cdot 1_{\left\{ \sigma_{t+1} > \sigma_t \right\}} \right) \log \left( \frac{ \|m_{t+1}\| }{ \|m_{t}\| } \right) 
	\\
	&+ v \cdot \log^+ \left( \frac{\alpha^{5/4} \cdot \elll \cdot \|m_{t}\|}{d \cdot \sigma_{t}} \right)  \cdot 1_{\left\{ \sigma_{t+1} < \sigma_t \right\}} \\
	&+ v \cdot \log^+ \left( \frac{\alpha^{5/4} \cdot d \cdot \sigma_{t}}{\uu \cdot \|m_{t}\|} \right) \cdot 1_{\left\{ \sigma_{t+1} > \sigma_t \right\}} 
	\\
	\leq& (1 - v )  \log \left( \frac{ \|m_{t+1}\| }{ \|m_{t}\| } \right) \\
	&+ v \cdot \log^+ \left( \frac{\alpha^{5/4} \cdot \elll \cdot \|m_{t}\|}{d \cdot \sigma_{t}} \right) \cdot 1_{\left\{ \sigma_{t+1} < \sigma_t \right\}} \\
	&+ v \cdot \log^+ \left( \frac{\alpha^{5/4} \cdot d \cdot \sigma_{t}}{\uu \cdot \|m_{t}\|} \right) \cdot 1_{\left\{ \sigma_{t+1} > \sigma_t \right\}} 
	\enspace.
\end{align*}
The truncated one-step change $\max\{ \Delta_{t} \, , \, - A\}$ is upper bounded by
\begin{align*}
	\max\{ \Delta_{t} \, , \, - A\} 
	\leq& (1 - v ) \max\left\{  \log \left( \frac{ \|m_{t+1}\| }{ \|m_{t}\| } \right)  \, , \, - \frac{A}{1 - v} \right\}\\
	&+ v \cdot \log^+ \left( \frac{\alpha^{5/4} \cdot \elll \cdot \|m_{t}\|}{d \cdot \sigma_{t}} \right) \cdot 1_{\left\{ \sigma_{t+1} < \sigma_t \right\}} \\
	&+ v \cdot \log^+ \left( \frac{\alpha^{5/4} \cdot d \cdot \sigma_{t}}{\uu \cdot \|m_{t}\|} \right) \cdot 1_{\left\{ \sigma_{t+1} > \sigma_t \right\}}  \enspace.
\end{align*}
To consider the expectation of the above upper bound, we need to compute
the expectation of the maximum of
$\log \left( \frac{ \|m_{t+1}\| }{ \|m_{t}\| } \right)$ and
$-\frac{A}{1 - v}$. Let $a \leq 0$ and $b \in \R$ then
\begin{align*}
	\max(a, b) = a \cdot \ind{a > b} + b \cdot \ind{a \leq b} \leq b \cdot \ind{a \leq b} \enspace.
\end{align*}
Applying this and taking the conditional expectation, a trivial upper
bound for the conditional expectation of
\begin{equation*}
\max\left\{ \log \left( \frac{ \|m_{t+1}\| }{ \|m_{t}\| } \right)  \, , \, - \frac{A}{1 - v} \right\} 
\end{equation*}
is $-\frac{A}{1 - v}$ times the probability of
$\log \left( \frac{ \|m_{t+1}\| }{ \|m_{t}\| } \right)$
being no greater than $-\frac{A}{1 - v}$.
The latter condition is equivalent to
$\|m_{t+1}\| \leq (1-r) \cdot \|m_{t}\|$ corresponding to successes with
rate $r = 1 - \exp\left( - \frac{A}{1 - v}\right)$ or better.
That is, 
\begin{multline}
	(1 - v ) \cdot \Expectation \left[ \max\left\{  \log \left( \frac{ \|m_{t+1}\| }{ \|m_{t}\| } \right)  \, , \, - \frac{A}{1 - v} \right\} \right] \\
	\leq - A \cdot p^\text{succ}_{r, d}\left(\frac{d \cdot \sigma_t}{\|m_t\|}\right) 
\end{multline}
Note also that the expected value of $\indlr{ \sigma_{t+1} > \sigma_t }$
is the success probability $p^\text{succ}_{0, d}\left(\frac{d \sigma_t}{\|m_t\|}\right)$.
We obtain an upper bound for the conditional expectation of
$\max\{ \Delta_{t} \, , \, - A\}$ in the case of reasonable $\sigma_t$
as
\begin{multline}
	\Expectation\left[\max\{ \Delta_{t} \, , \, - A\} | \theta_t \right] \leq  - A \cdot p^\text{succ}_{r, d}\left(\frac{d \sigma_t}{\|m_t\|}\right)  \\
	+ \left( \frac54 \log (\alpha) + \underbrace{ \log\left( \frac{ \elll \| m_t \|}{d \sigma_t} \right)}_{\leq 0} \right) \cdot v \cdot \left( 1 - p^\text{succ}_{0, d}\left(\frac{d \sigma_t}{\|m_t\|}\right) \right) \\
	+ \left( \frac54 \log (\alpha) + \underbrace{\log \left( \frac{d \sigma_t}{\uu \| m_t \|} \right)}_{\leq 0} \right)  \cdot v \cdot p^\text{succ}_{0, d}\left(\frac{d \sigma_t}{\|m_t\|}\right) \\
	\leq - A \cdot p^* + \frac54 \log (\alpha) \cdot v
	\enspace, \qquad\qquad
	\label{eq:case3bound}
\end{multline}
where $r = 1 - \exp\left( - \frac{A}{1 - v}\right)$.

\textit{Small $\sigma_t$ case: $\frac{\|m_t\|}{d \sigma_t} > \frac{1}{\elll}$}.
If $\ell \|m_t\| > d \sigma_t$, the summand~\eqref{eq:a} is positive.
Moreover, if $\sigma_{t + 1} < \sigma_{t}$, we have
$\elll \|m_{t+1}\| / d = \elll \|m_{t}\|/d > \sigma_t > \sigma_{t+1}$ and hence
the summand~\eqref{eq:a} is positive for $V(\theta_{t+1})$ as well. If
$\sigma_{t + 1} > \sigma_{t}$, any regime can happen. Then,
\begin{align*}
	&V(\theta_{t+1}) - V(\theta_{t})\notag \\
	&= \log \left( \frac{\|m_{t+1}\|}{ \|m_{t}\| } \right) - v \cdot \log \left( \frac{\alpha \cdot \elll \cdot \|m_t\|}{d \cdot \sigma_t} \right)\notag \\
	&+ v \cdot \log \left( \frac{\alpha \cdot \elll \cdot \|m_{t+1}\|}{d \sigma_{t+1}} \right) \indlr{ \alpha \elll \|m_{t+1}\| > d \cdot \sigma_{t+1} }\indlr{ \sigma_{t+1} > \sigma_{t} } \notag \\
	&+ v \cdot \log \left( \frac{\alpha \cdot \elll \cdot \|m_{t+1}\|}{d \sigma_{t+1}} \right) \indlr{ \alpha \elll \|m_{t+1}\| > d \cdot \sigma_{t+1} }\indlr{ \sigma_{t+1} < \sigma_{t} } \notag \\
	&+ v \cdot \log \left( \frac{\alpha^{1/4} \cdot \sigma_{t+1}d}{\uu \cdot \|m_{t+1}\|} \right) \indlr{\alpha^{-1/4} \uu \|m_{t+1}\| < d \cdot \sigma_{t+1} } \indlr{ \sigma_{t+1} > \sigma_{t} } \notag 
\end{align*}	
\begin{align*}	
	&= \Big[1 + \left(v \cdot \indlr{ \elll \|m_{t+1}\| > d \sigma_{t} } - v \cdot \indlr{\alpha^{-5/4} \uu \|m_{t+1}\| < d \cdot \sigma_{t} } \right) \notag \\
	& \qquad \cdot \indlr{ \sigma_{t+1} > \sigma_{t} } \Big] \cdot \log \left( \frac{\|m_{t+1}\|}{ \|m_{t}\| } \right)\notag \\
	&- v \cdot \log \left( \frac{\uu \|m_{t}\|}{\alpha^{5/4} \cdot d \cdot \sigma_{t}} \right) \indlr{\alpha^{-5/4} \uu \|m_{t+1}\| < d \cdot \sigma_{t} } \indlr{ \sigma_{t+1} > \sigma_{t} } \notag \\
	&- v \cdot \log \left( \frac{\elll \|m_{t}\|}{d \cdot \sigma_t} \right) \cdot \Big( 1 - \indlr{ \elll \|m_{t+1}\| > d \cdot \sigma_{t} }\indlr{ \sigma_{t+1} > \sigma_{t} } \notag \\
	& \qquad - \indlr{ \alpha^{5/4} \elll \|m_{t+1}\| > d \cdot \sigma_{t} }\indlr{ \sigma_{t+1} < \sigma_{t} } \Big) \notag \\
	&- v \cdot \log ( \alpha) \cdot \left( 1 - \frac54 \indlr{ \alpha^{5/4} \elll \|m_{t+1}\| > d \cdot \sigma_{t} }\indlr{ \sigma_{t+1} < \sigma_{t} } \right) \label{eq:case1}
\end{align*}
On the RHS of the above equality, the first term is guaranteed to be
non-positive since $v \in (0, 1)$.
The second and third terms are non-positive as well since
$\frac{\uu \|m_{t}\|}{d \alpha^{5/4} \sigma_{t}} > \frac{\uu}{\alpha^{5/4} \elll} > 1$
and $\frac{\elll \|m_{t}\|}{d \sigma_{t}} > 1$. Since
$v \cdot \log(\alpha)$ is positive, replacing the indicator
$\indlr{ \alpha^{5/4} \elll \|m_{t+1}\| > d \sigma_{t} }$ with
$1$ provides an upper bound. Altogether, we obtain
\begin{equation*}
	V(\theta_{t+1}) - V(\theta_{t})
	\leq - v \cdot \log ( \alpha) \cdot \left( 1 - \frac54 \indlr{ \sigma_{t+1} < \sigma_{t} } \right)  \enspace.
\end{equation*}
Note that the RHS is larger than $- A$. Then, the conditional
expectation of $\max\{ \Delta_{t} \, , \, - A\}$ is
\begin{align}
  \Expectation\left[\max\{ \Delta_{t} \, , \, - A\} | \mathcal{F}_t \right]
  & \leq - v \cdot \log ( \alpha) \cdot \left( \frac54 p^\text{succ}_{0,d}\left( \frac{d \sigma_t}{\|m_t\|} \right) - \frac14 \right)  \notag\\
  & \leq - v \cdot \log ( \alpha) \cdot \left( \frac{5 p_{\ell} - 1}{4} \right) < 0 \enspace.\label{eq:case1bound}
\end{align}
Here we used $p^\text{succ}_{0,d}\left( \frac{d \sigma_t}{\|m_t\|} \right) > p_\ell > 1/5$.

\textit{Large $\sigma_t$ case: $\frac{\|m_t\|}{d \sigma_t} < \frac{1}{\uu}$}.
Since $\frac{\|m_{t+1}\|}{\sigma_{t+1}} \leq \frac{\|m_t\|}{\alpha^{-1/4} \sigma_t} < \frac{d}{\alpha^{-1/4} \uu}$,
the summand~\eqref{eq:b} is positive in both $V(\theta_t)$ and $V(\theta_{t+1})$.
For the summand~\eqref{eq:a}, recall that $\alpha \elll \| m_t \|/ d\sigma_t < \alpha \elll / \uu \leq \alpha \cdot \alpha^{-5/4} = \alpha^{-1/4} < 1$
since we have assumed that $\uu / \elll \geq \alpha^{5/4}$. Hence, for
$V(\theta_t)$ the summand~\eqref{eq:a} is zero. Also,
$\alpha \elll \| m_{t+1} \| / d \sigma_{t+1} \leq \alpha \elll / (\alpha^{-1/4} \uu ) = \alpha^{5/4} \elll / \uu \geq 1$
and thus for $V(\theta_{t+1})$ the summand~\eqref{eq:a} also equals $0$.
We obtain
\begin{align*}
	V(\theta_{t+1}) - V(\theta_t) 
	= & (1 - v) \Big( \log\left( \| m_{t+1} \| \right) - \log\left( \| m_t \| \right) \Big) \\
	  & + v \cdot \log\left( \sigma_{t+1} / \sigma_t \right),
\end{align*}
where $\log\left( \sigma_{t+1} / \sigma_t \right)$ equals
$\log(\alpha)$ with probability
$p^\text{succ}_{0,d}\left( \frac{d \sigma_t}{\|m_t\|} \right)$,
and $-\frac{1}{4} \log(\alpha)$ with probability
$1 - p^\text{succ}_{0,d}\left( \frac{d \sigma_t}{\|m_t\|} \right)$.
The first term on the RHS is guaranteed to be non-positive since $v < 1$,
yielding $\Delta_{t} \leq v \cdot \log(\sigma_{t+1}/\sigma_t)$.
On the other hand,
\begin{align*}
	& v \cdot \log(\sigma_{t+1}/\sigma_t)   \\
	& =v \cdot \left( \log (\alpha)  \ind{ \| m_{t+1}\| < \| m_t \| \}} - \frac14 \log(\alpha) \ind{ \| m_{t+1}\| = \| m_t \| \}} \right) \\
	& =v \cdot \left( \frac54 \log(\alpha)  \ind{ \| m_{t+1}\| < \| m_t \| \}} - \frac14 \log(\alpha) \right) \\
	& \geq - \frac14 \log(\alpha) v \geq -A
\end{align*}
where the last inequality comes from the prerequisite \erratum{$v < A / \log(\alpha)$}. 
Hence,
$\max \{ v \cdot \log(\sigma_{t+1}/\sigma_t) , - A\} = v \log(\sigma_{t+1} / \sigma_t)$
such that $\max\{ \Delta_{t} \, ,\allowbreak \, - A\} \leq v \cdot  \log(\sigma_{t+1} / \sigma_t)$.
Then, the conditional expectation of $\max\{ \Delta_{t} \, ,\allowbreak \, - A\}$ is
\begin{align}
	\Expectation\left[\max\{ \Delta_{t} \, , \, - A\} | \theta_t \right]
	& \leq - \frac14 v \log(\alpha)\left(1 - 5  p^\text{succ}_{0,d}\left( \frac{d \sigma_t}{\|m_t\|} \right) \right) \notag\\
	& \leq - v \log(\alpha)\left(\frac{1 - 5  p_{u}}{4} \right) < 0 \enspace.
	\label{eq:case2bound}
\end{align}
Here we used $p^\text{succ}_{0,d}\left( \frac{d \sigma_t}{\|m_t\|} \right) \leq p_{u} < 1/5$.

Inequalities \eqref{eq:case1bound}, \eqref{eq:case2bound}, and \eqref{eq:case3bound}
together cover all possible cases and hence imply the bound~\eqref{eq:B}.
\end{proof}
\hrule
\begin{proof}[proof of proposition \ref{proposition:scaling}]
We rewrite $r' = 1 - \exp\left(-\frac{A}{1 - \frac{1}{d \cdot \log(\alpha)}}\right)$.
It holds $v < \frac{1}{d \cdot \log(\alpha)}$ and hence
$r' > r$, from which we obtain $p' < p^*$.
Now we consider the terms in equation~\eqref{eq:B} one by one.
We start with
\begin{align*}
	A \cdot p^* - \frac54 \cdot \log(\alpha) \cdot v
	= \frac{p^*}{d} - \frac58 \frac{p'}{d}
\end{align*}
which is lower bounded by $-\frac{3}{8} \frac{p'}{d}$
and upper bounded by $-\frac{3}{8} \frac{p^*}{d}$.
Furthermore, we obtain
\begin{align*}
	& \log(\alpha) \cdot v \cdot \frac{5 p_\ell - 1}{4} = \frac{p'}{d} \cdot \frac{5 p_\ell - 1}{8} \\
	& \log(\alpha) \cdot v \cdot \frac{1 - 5 p_u}{4} = \frac{p'}{d} \cdot \frac{1 - 5 p_u}{8}
	\enspace.
\end{align*}
We collect these results into the definition of lower and upper bounds
\begin{align*}
	L &= \frac{p'}{d} \cdot \min \left\{
			\frac38, \frac{5 p_\ell - 1}{8}, \frac{1 - 5 p_u}{8}
		\right\} \\
	U &= \frac{ p^* }{d} \cdot \max \left\{
			\frac38, \frac{5 p_\ell - 1}{8}, \frac{1 - 5 p_u}{8}
		\right\}
\end{align*}
for $L \leq B \leq U$. From $L > 0$ we immediately obtain $B > 0$.
We have $\lim_{d \to \infty} d \cdot r = 1$ and hence according to
Lemma~\ref{lemma:rho}
\begin{align*}
	 & \lim_{d \to \infty} p^*
		= \lim_{d \to \infty} \left( \min\limits_{\barsigma \in [\elll, \uu]} \left\{ p^\text{succ}_{r, d} (\barsigma) \right\} \right) \\
		& \overset{(\star)}{=} \min\limits_{\barsigma \in [\elll, \uu]} \left\{ \lim_{d \to \infty} \left( p^\text{succ}_{r, d} (\barsigma) \right) \right\}
		  = \min\limits_{\barsigma \in [\elll, \uu]} \left\{ \Phi_1\left(-\frac{1}{\barsigma} - \frac{{\barsigma}}{2}\right) \right\} \\
		& = \min \left\{ \Phi_1\left(-\frac{1}{{\elll}} - \frac{{\elll}}{2}\right), \Phi_1\left(-\frac{1}{{\uu}} - \frac{{\uu}}{2}\right) \right\}
		  > 0
	\enspace.
\end{align*}
The equality $(\star)$ holds as follows: Let $(\barsigma_d)_{d \in \mathbb{N}}$ be a
sequence of points where the minimum is attained, then the Bolzano-Weierstra{\ss}
property provides a convergent sub-sequence with limit point $\barsigma \in [\elll, \uu]$.
Since the success probability functions and its limit are continuous, the minimum
of the limit function is attained at $\barsigma$.
We obtain $U \in \Theta(1/d)$. Analogously, with $\lim_{d \to \infty} d \cdot r' = 1$
and
\begin{align*}
	\lim_{d \to \infty} p'
		= \min\limits_{\barsigma \in [\elll, \uu]} \left\{ \Phi_1\left(-\frac{1}{\barsigma} - \frac{{\barsigma}}{2}\right) \right\}
		> 0
\end{align*}
we also obtain $L \in \Theta(1/d)$. Combining the results for $L$ and
$U$ proves $B \in \Theta(1/d)$.
\end{proof}
\hrule
\begin{proof}[proof of lemma \ref{lemma:hitandrun}]
The log progress of the hit-and-run algorithm amounts to
$-\log( \sin(\theta)) \cdot 1_{\{\theta \leq \pi/2\}}$, where
$\theta \in [0, \pi)$ is the angle between $\delta_t$ and $e_1$. This
follows from a geometric interpretation of the algorithm. Let
$W_d = \int_0^{\pi/2} \sin^d(\theta) d \theta$ denote the Wallis
integral. Then the density of $\theta$ is
$(2 W_{d-2})^{-1} |\sin(\theta)|^{d-2}$. The expected log progress of
the hit-and-run algorithm is written as
\begin{multline*}
\frac{-1}{2 W_{d-2}} \int_{0}^{\pi/2} \log \left( \sin(\theta)  \right) \sin^{d-2}(\theta) d\theta \\
= - \big(2 (d-1)^2 W_{d-2} \big)^{-1} \int_{0}^{1} \log \left( r \right) /  \sqrt{ 1 - r^{2 / (d- 1)}} d r \enspace.
\end{multline*}
Here we applied the change of variables $\sin(\theta)^{d-1} = r$. When
considering $r$ as a random variable uniformly distributed on $[0, 1]$,
then $\log \left( r \right)$ and $1/ \sqrt{ 1 - r^{2 / (d- 1)}}$ are
positively correlated \cite[Chapter~1, eq.~(2.1)]{thorisson2000coupling}.
Therefore, the integral on the RHS is lower bounded by the product of
the integrals of the two terms, which reads
\begin{align*}
\int_{0}^{1} \frac{\log \left( r \right)}{\sqrt{1 - r^{2 / (d- 1)}}} d r
  \geq \int_{0}^{1} \log \left( r \right)d r \int_{0}^{1} \sqrt{\frac{1}{1 - r^{2 / (d- 1)}}} d r
\enspace,
\end{align*}
where the 1st and 2nd integral on the RHS are $-1$ and $(d - 1)
W_{d-2}$, respectively. Using $d \leq 2(d-1)$ for all $d \geq 2$
concludes the proof.
\end{proof}
\hrule
\begin{proof}[proof of theorem \ref{theorem:main}]
Since $\log(\|m_t\|) \leq V(\theta_t)$, then the hitting time of
$(-\infty,\allowbreak \log(\epsilon)]$ by $V(\theta_t)$, denoted $T_\epsilon^V$,
is not less than $T_\epsilon$. In Proposition~\ref{proposition:drift} we have
shown that for $A=1/d$ and $v$ as set in Proposition~\ref{proposition:scaling},
the drift
\begin{equation*}
	\Expectation\left[ \max\{ V(\theta_{t+1}) - V(\theta_{t}) \, ,\, -\mathrm{1/d} \} \mid \theta_{t}\right] \leq - B
\end{equation*}
holds.
By applying Theorem~\ref{theo:drift-UB-trunc} we obtain
$$
	\E \left[T_\epsilon^V \right]
	\leq \big( V(\theta_0) - \log(\epsilon) + 1/d \big) \big/ B        
	\in \Theta \Big( (V(\theta_0) - \log(\epsilon)) \cdot d \Big)
	\enspace.
$$
Together with $\E \left[T_\epsilon \right] \leq \E \left[T_\epsilon^V \right]$
this shows the upper bound.
Lemma \ref{lemma:hitandrun} bounds the drift of $X_t = \log(\|m_t\|)$,
see eq.\ \eqref{LB-drift}. With the bound $C = \frac{1}{d}$ and
$\target = \log(\epsilon)$, Theorem \ref{theorem:drift-anne-jens} yields
the lower bound
$\E[T_\epsilon] \geq \frac{(x_0 - \target) \cdot d}{4} - \frac{1}{2} \in \Theta \Big( \big(x_0 - \log(\epsilon)\big) \cdot d \Big)$.
\end{proof}

\end{multicols}

\end{document}